\documentclass{article}

\usepackage{graphicx}
\usepackage{wrapfig}
\usepackage{amsmath}
\usepackage[percent]{overpic}
\usepackage{xcolor}
\usepackage{subcaption}
\usepackage{siunitx}
\usepackage{etoolbox}

\usepackage[nonatbib, final]{neurips_2020}

\usepackage[utf8]{inputenc} 
\usepackage[T1]{fontenc}    
\usepackage{hyperref}       
\usepackage{url}            
\usepackage{booktabs}       
\usepackage{amsfonts}       
\usepackage{nicefrac}       
\usepackage{microtype}      

\title{Spring-Rod System Identification via Differentiable Physics Engine}

\author{%
    Kun Wang,  Mridul Aanjaneya and Kostas Bekris\\
    Rutgers University \\
    \texttt{\{kun.wang2012, mridul.aanjaneya, kostas.bekris\}@rutgers.edu}\\
 
}

\begin{document}
\vspace{-100mm}
\maketitle
\begin{abstract}
\vspace{-4mm}
We propose a novel differentiable physics engine for system identification of complex spring-rod assemblies. Unlike black-box  data-driven methods for learning the evolution of a dynamical system \emph{and} its parameters, we modularize the design of our engine using a discrete form of the governing equations of motion, similar to a traditional physics engine. We further reduce the dimension from 3D to 1D for each module, which allows efficient learning of system parameters using linear regression. The regression parameters correspond to physical quantities, such as spring stiffness or the mass of the rod, making the pipeline explainable. The approach significantly reduces the amount of training data required, and also avoids iterative identification of data sampling and model training. We compare the performance of the proposed engine with previous solutions, and demonstrate its efficacy on tensegrity systems, such as NASA's icosahedron.
\end{abstract}

\section{Introduction}
\vspace{-.1in}
Cable-driven robots are gaining increasing attention due to their adaptiveness and safety. Tensegrity structures have many applications: from manipulation \cite{lessard2016bio}, locomotion \cite{sabelhaus2018design}, morphing airfoil \cite{chen2020design} to spacecraft lander \cite{bruce2014superball}. While useful and versatile, they are difficult to accurately model and control. Identifying system parameters is necessary, either to learn controllers in simulation (as real-world experiments are time-consuming, expensive and dangerous), or for traditional model-based control. In all these cases, the spring-rod representation considered in this work is the basic modeling element. 

However, the spring-rod system has high degrees of freedom for system identification. Physics-based methods for simulation require accurate models that capture non-linear material behavior, which are difficult to construct. In contrast, data-driven methods can simulate any system from observed data, with sufficient training data. But the large number of variables and non-linear material properties necessitate copious amounts of training data.

Motivated by these issues, we propose a data-driven differentiable physics engine that combines the benefits of data-driven and physics-based models, while alleviating most of their drawbacks, and is designed from first principles. Previous data-driven models have required large amounts of data, because they learn the  parameters \emph{and} the physics of the system.  Furthermore, the hidden variables and black box nature of these models are not explainable, and difficult to transfer to new environments. Our approach is based on the observation that the equations that govern the motion of such systems are well-understood, and can be directly baked into the data-driven model. Such a design can reduce demands on training data and can also generalize to new environments, as the governing principles remain the same. We further simplify the differentiable engine by modularization, which compartmentalizes the problem of learning the dynamics of the whole system to smaller well-contained problems. For each module, we also reduce the dimension from 3D to 1D, by taking advantage the properties of spring-rod systems, which allows for efficient parameter inference using linear regression. As a side benefit, the regression parameters correspond to physical quantities, such as the spring stiffness or the mass of the rod, making the framework explainable.

\begin{figure}
\vspace{-0.7cm}
    \centering
    \includegraphics[height=0.25\linewidth, trim={6.5cm 5cm 2cm 1cm}, clip]{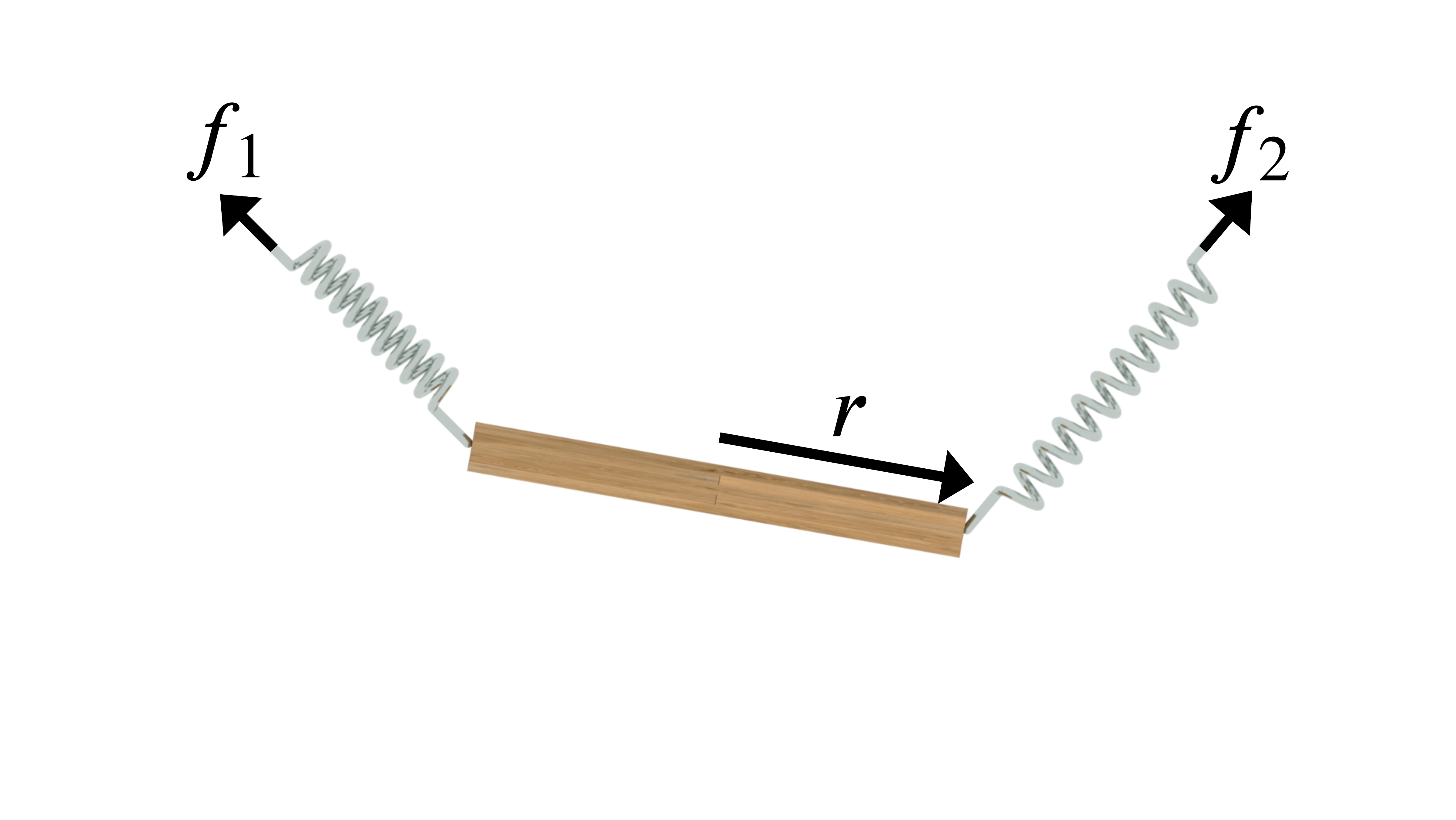}
    \includegraphics[height=0.25\linewidth, trim={0cm 0cm 0cm 0cm}, clip]{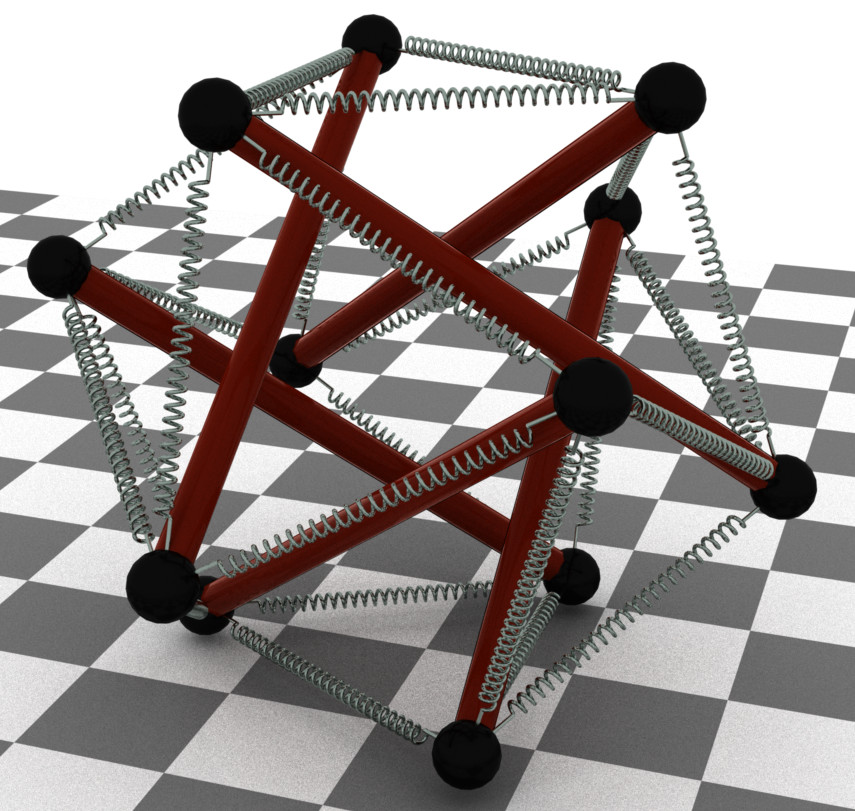}
    \includegraphics[height=0.25\linewidth]{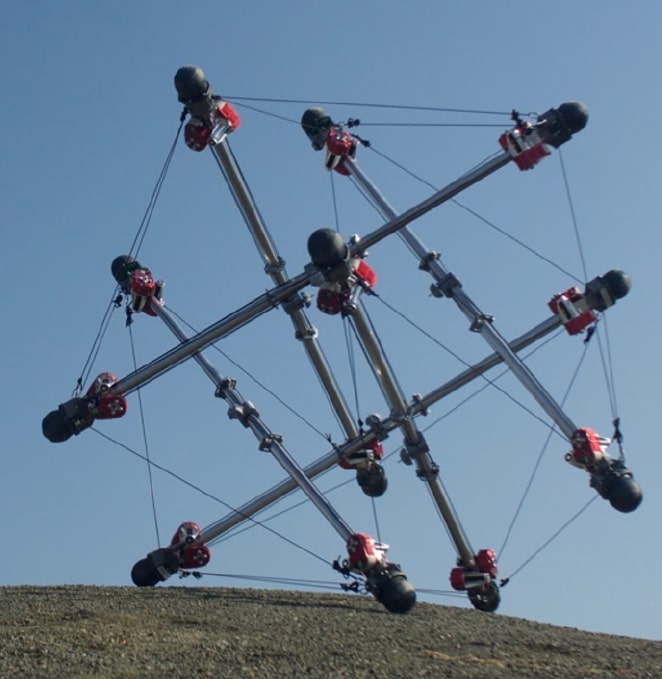}
    \vspace{-2mm}
    \caption{A basic element with one rod connected by two springs (left). A complex assembly of rods and springs forming a tensegrity robot in simulation (middle), and the real world (right).}
    \label{fig:spring_rod_model}
\end{figure}

\vspace{-3mm}
\section{Related Work}
\vspace{-.1in}
Traditional methods for system identification build a dynamics model by minimizing the prediction error~\cite{swevers1997optimal} \cite{hansen2001completely}.
These methods require parameter refinement and data sampling in an iterative fashion, to decrease the prediction error. This iterative process can be avoided using data-driven techniques that directly fit a physics model to data~\cite{rosenblatt1958perceptron,rumelhart1986learning,asenov2019vid2param}. However, these techniques treat the dynamics as a black box, are data hungry, and require retraining in a new environment.

Instead of treating the environment as a black box, \emph{interaction network} \cite{battaglia2016interaction} took the first step to modularize objects and their interactions. Later, researchers extended this idea by a hierarchical relation network for graph-based object representation~\cite{mrowca2018flexible} and a propagation network \cite{li2019propagation} for the multi-step propagation network. While these methods are an improvement over previous approaches, they still treat the interactions between different objects as black boxes and try to learn them from data, even though the governing equations of motion are well-understood.

Quite a few authors have recently introduced differentiable physics engines that focus on many aspects not central to our work. For example, forward dynamics ~\cite{heiden2019interactive}, Material Point Method (MPM) \cite{hu2019chainqueen}, linear complementarity problems (LCP)~\cite{de2018end} , augmented Lagrangian~\cite{landry2019differentiable}, differentiable programming~\cite{hu2019difftaichi}, augmented traditional physics simulators\cite{sain}, and LSTM dynamics model without system topology~\cite{golemo2018sim}. Researchers have also proposed differentiable engines specific to certain kinds of objects, such as molecules~\cite{jaxmd2019}, fluids~\cite{spnets2018}, and cloth~\cite{liang2019differentiable}. Recent works on tensegrity robots~\cite{surovik2019adaptive, littlefield2019kinodynamic, surovid2018any} make major improvement on locomotion in simulation and have great challenges on policy transfer to real world system. All these works motivate us to mitigate the reality gap for cable driven robots between simulation and reality via system identification.

\vspace{-3mm}
\section{Methods}
\vspace{-0.1in}

\begin{figure} 
\vspace{-0.3cm}
    \centering
    \includegraphics[width=\textwidth]{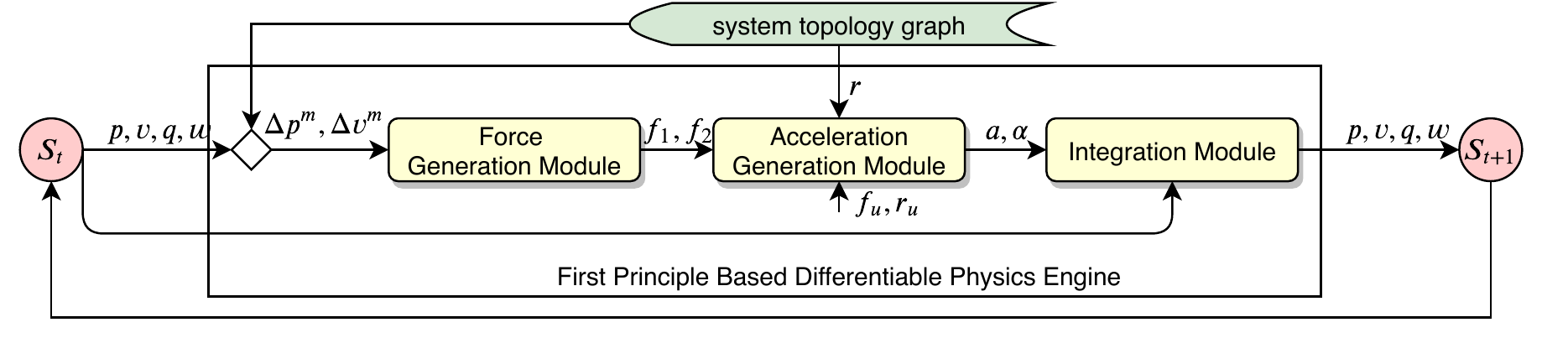}
    \vspace{-7mm}
    \caption{Flow chart showing the data flow when simulating one time step with our physics engine.}
    \label{fig:architecture}
    \vspace{-5mm}
\end{figure}

Our system views a spring-rod system as a composition of basic \emph{elements} (see Fig.~\ref{fig:spring_rod_model}(left)), where springs generate forces that influence rod dynamics. We subdivide each time step of the simulation into three modules: force generation, acceleration computation, and state update/integration (see Fig.~\ref{fig:architecture}). The physics engine takes as input the current rod state $S_t = \{\boldsymbol{p, v, q, \omega}\}$, where $p$ is position, $v$ is linear velocity, $q$ is orientation (expressed as a quaternion), and $\omega$ is the angular velocity. Based on $S_t$, the position and linear velocity $p^{m}, v^{m}$ of the two rod endpoints is computed, and is used to compute the relative compression (or expansion) and velocity $\Delta p^{m}, \Delta v^{m}$ of the two attached springs. Then, the first module predicts the spring forces $f$, the second module computes the linear and angular accelerations $a,\alpha$, and the third module integrates the new state $S_{t+1}$.


\vspace{-2mm}
\subsection{System Topology Graph}
\vspace{-2mm}

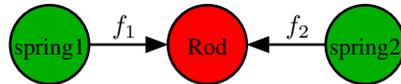
\begin{wrapfigure}[4]{r}{0.38\textwidth} 
\vspace*{-6mm}
\begin{overpic}[width=0.38\textwidth]{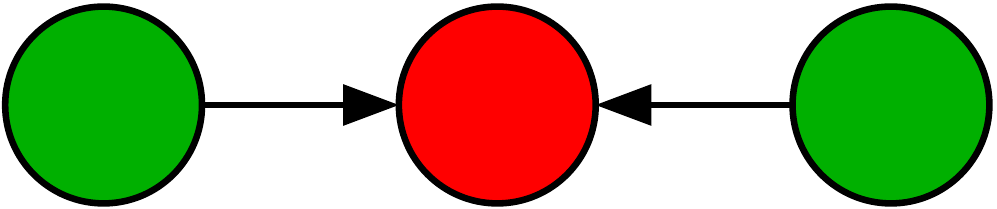}
    \put(1.5,8){{\small spring1}}
    \put(26,13){{$f_1$}}
    \put(45,8){{\small Rod}}
    \put(69.5,13){{$f_2$}}
    \put(80.5,8){{\small spring2}}
\end{overpic}
\vspace{-5mm}
\caption{\small Element topology graph.}
\label{fig:topology}
\end{wrapfigure}

We use a topology graph to represent interconnections between different components of the spring-rod system. Each rod and spring has a corresponding vertex, and directed edges represent relations between them. Figure~\ref{fig:topology} shows an example topology graph for the basic spring-rod element shown in Figure~\ref{fig:spring_rod_model}(left).

\vspace{-3mm}
\subsection{Force Generation Module}
\vspace{-3mm}

\begin{figure}
    \centering
    \includegraphics[width=0.8\textwidth]{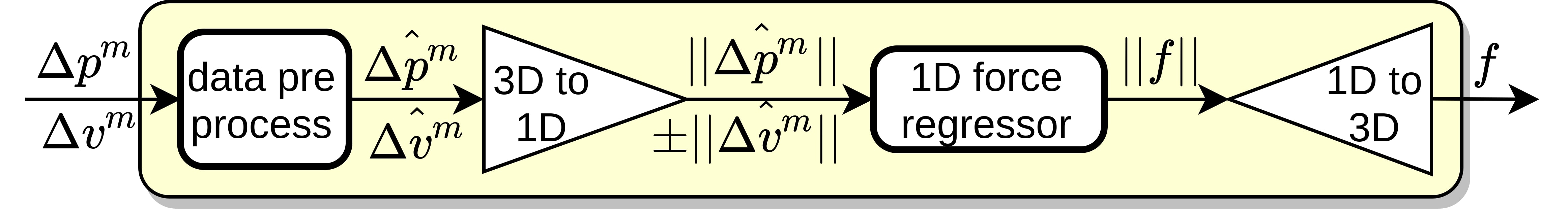}
    \vspace{-1mm}
    \caption{Force generation module, which uses dimensionality reduction to compute spring forces.}
    \label{fig:spring_physics_engine}
 \vspace{-5mm}
\end{figure}

The relative compression (or expansion) $\Delta p^{m}$ and velocity $\Delta v^{m}$ of each spring is given as input to the force generation module, which outputs the spring forces $f$  by Hooke's law. As shown in Fig.\ref{fig:spring_physics_engine}, two unknown parameters, stiffness $K$ and damping $c$, can be easily learned using linear regression.

\vspace{-3mm}
\subsection{Acceleration Generation Module}
\vspace{-3mm}

\begin{figure}[t] 
    \centering
    \includegraphics[width=0.9\textwidth]{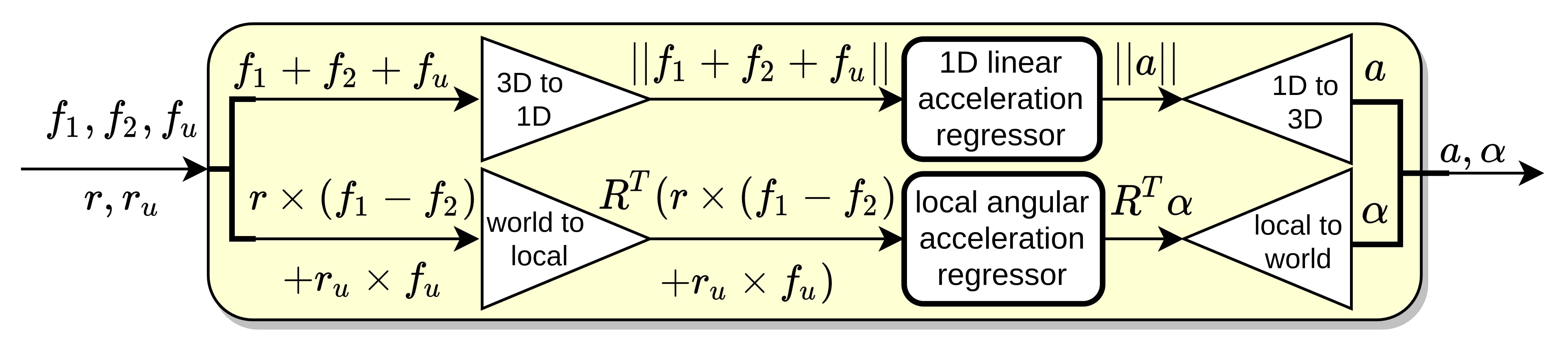}
    \vspace{-2mm}
    \caption{Acceleration generation module}
    \label{fig:rod_physics_engine}
    \vspace{-6mm}
\end{figure}
The acceleration generation module takes the spring forces $f$ and control force $f_u$ as input and outputs each rod's linear and angular accelerations $a,\alpha$ as shown in Fig~\ref{fig:rod_physics_engine}. $f_1$ and $f_2$ are spring forces on the two rod ends, $f_u$ is control force, $r$ is the half-length rod vector, $r_u$ is control force arm, $R$ is the rod local/world frame rotation matrix. The rod mass $M$ and inertia $I$ are unknown parameters to identify.

\vspace{-3mm}
\subsection{Integration Module and Method Implementation}
\label{integraion_module}
\vspace{-3mm}

The integration module computes forward dynamics of each rod using the current accelerations $a,\alpha$. We apply the semi-implicit Euler method~\cite{stewart2000implicit} to compute the updated state $S_{t+1}=\{p_{t+1}, v_{t+1}, q_{t+1}, \omega_{t+1}\}$ at the end of the current time step.

The learning module receives the current state $S_t$ and returns a prediction $\hat{S_{t+1}}$.  The loss function is the MSE between the predicted $\hat{S_{t+1}}$ and ground truth state $S_{t+1}$. The proposed decomposition, first-principles approach and the cable’s linear nature allow the application of linear regression, which helps with data efficiency. This linear regression step has been implemented as a single layer neural network without activation function on pyTorch \cite{paszke2019pytorch}.

\vspace{-3mm}
\section{Experiments}
\vspace{-3mm}
The task is to identify parameters including spring stiffness $K$, damping $c$ and rod mass $M$, inertia $I$.
\vspace{-3mm}
\subsection{Simple Spring-Rod Model Identification}

\begin{figure}
\vspace{-0.3cm}
    \centering
    \begin{minipage}[b]{.32\linewidth}
    \includegraphics[height=1.4cm, trim={1.8cm 0cm 0cm 0cm}, clip]{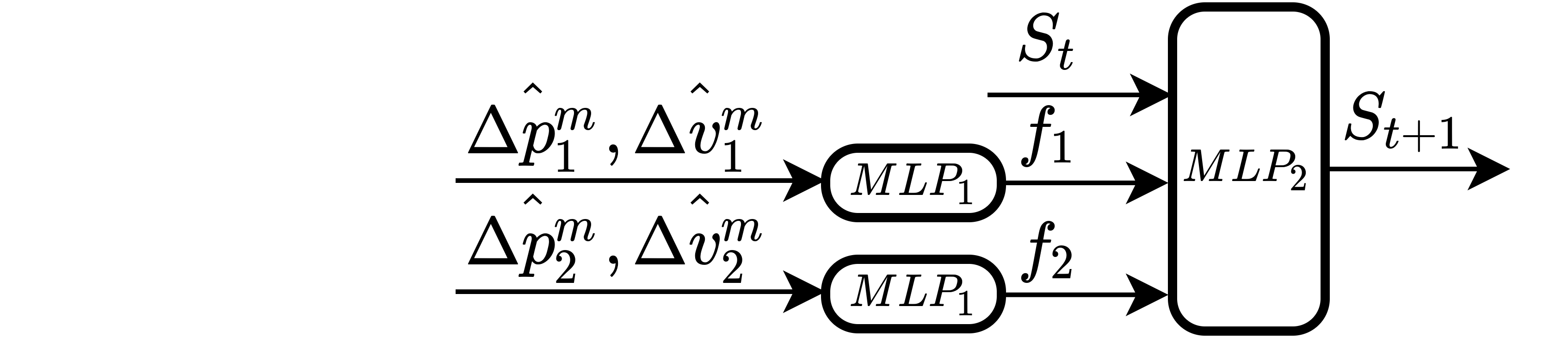}
    \vspace{-6mm}
    \caption{Interaction Network}
    \label{fig:interaction_network_architecture}
    \end{minipage}%
    \hfill 
    \begin{minipage}[b]{.67\linewidth}
    \includegraphics[width=\textwidth]{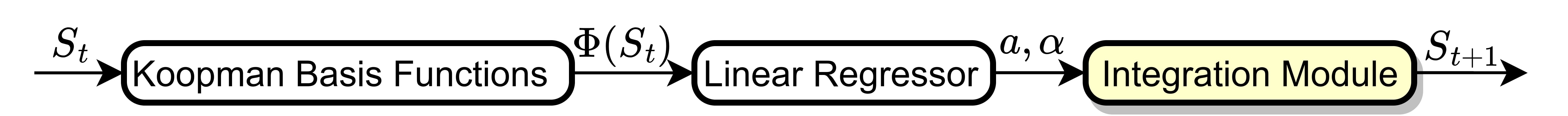}
    \vspace{-6mm}
    \caption{Koopman with Integration Module}
    \label{fig:koopman_w_int_architecture}
    \label{fig:interaction_network_w_int_architecture}
    \end{minipage}
\vspace{-0.2in}
\end{figure}
\begin{figure}
    \begin{minipage}[t]{0.5\linewidth}
    \centering
    \includegraphics[height=0.41\linewidth]{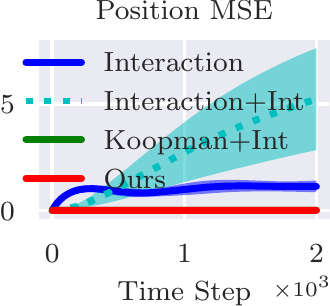}
    \includegraphics[height=0.41\linewidth]{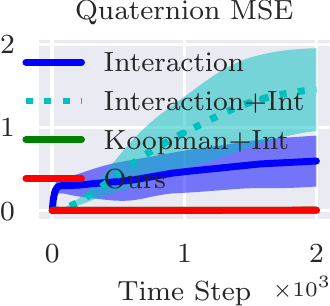}
    \caption*{a) Simple Spring-Rod Model Identification}
    \end{minipage}
    \begin{minipage}[t]{0.5\linewidth}
    \centering
    \includegraphics[height=0.41\linewidth]{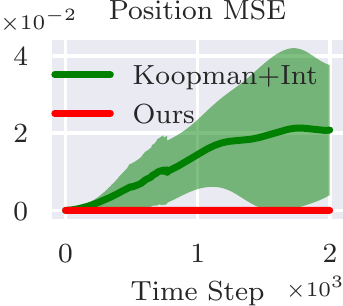}
    \includegraphics[height=0.41\linewidth]{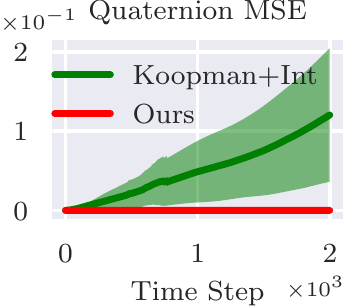}
    \caption*{b) Complex Tensegrity Model Identification}
    \end{minipage}
    \caption{Physical Parameters Estimation}
    \label{fig:superball_system_identification}
    \vspace{-0.6cm}
\end{figure}

First, we identify these parameters in a simple spring-rod system as shown in Figure~\ref{fig:spring_rod_model} (a). \textbf{Interaction} is an improved version of the Interaction Network \cite{battaglia2016interaction} as shown in Fig. \ref{fig:interaction_network_architecture}. It has two Multilayer Perceptrons (MLPs), one to generate spring forces $f$ and the other to generate rod state $S_{t+1}$. Unlike \cite{battaglia2016interaction}, which takes raw state $S_t$ as input, we generate $\hat{\Delta p^{m}_t}, \hat{\Delta v^{m}_t}$ as input. \textbf{Interaction+Int} appends the integration module to the Interaction Network, and replaces input $S_t$ of $MLP_2$ by $r$.  \textbf{Koopman+Int} is to use the Koopman operator to predict accelerations and apply the Integration Module to map them to $S_{t+1}$.  \textbf{Interaction} only predicts a $S_{t+1}$ in training data that is close to $S_t$. \textbf{Interaction+Int} experiences increasing error from accumulated prediction errors. The Koopman operator \textbf{Koopman+Int} designed from first principles gives accurate predictions similar to \textbf{Ours} in this simple system. Comparison of errors is shown in Figure~\ref{fig:superball_system_identification} a).

\vspace{-.1in}
\subsection{Complex Tensegrity Model Identification} 
\vspace{-.1in}
We consider an icosahedron tensegrity system as shown in Fig. \ref{fig:spring_rod_model} (c). It is composed of 6 rods and 24 springs. Each rod is connected to 8 springs and has a length of 1.04m. Each spring's rest length is 0.637m. We set the gravity constant to $g=-9.81$ in Mujoco. We collect 1000 trajectories with different initial conditions for training, 200 for validation and 100 for testing. The result is shown in Figure \ref{fig:superball_system_identification} b). \textbf{Our} approach outperforms \textbf{Koopman+Int} because designing basis functions for The Koopman operator has an increased data requirement relative to our approach. 

\vspace{-.1in}
\subsection{Data Efficiency Experiment}
\begin{figure}[ht]
\vspace{-.1in}
\begin{minipage}[t]{.5\linewidth}
\vspace{0pt}
\centering
\includegraphics[width=0.49\linewidth]{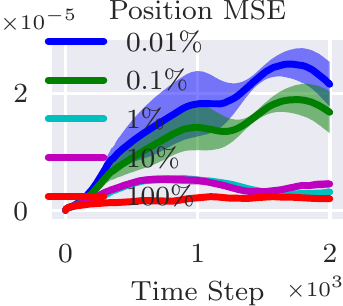}
\includegraphics[width=0.49\linewidth]{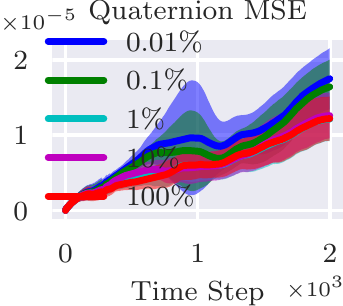}
\vspace{-0.5cm}
\caption*{a) Efficiency on Various Training Set}
\end{minipage}%
\begin{minipage}[t]{.5\linewidth}
\vspace{0pt}
\centering
\includegraphics[width=0.49\linewidth]{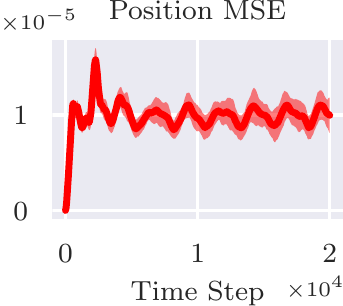}
\includegraphics[width=0.49\linewidth]{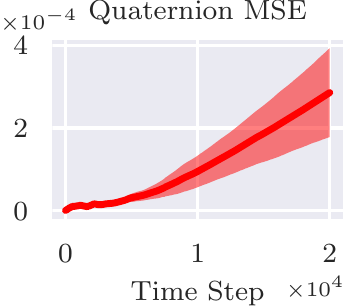}
\vspace{-0.5cm}
\caption*{b) Generalization on New Testing Set}
\end{minipage}
\vspace{-0.2cm}
\caption{Data Efficiency and Model Generalization Experiment.}
\label{fig:superball_diff_params_system_data_efficient_and_generalization}
\vspace{-0.15in}
\end{figure}

\vspace{-.1in}
The proposed method has relatively small data requirements as shown in Fig. \ref{fig:superball_diff_params_system_data_efficient_and_generalization} a). Instead of training on 1000 trajectories, which have 736,167 time steps in total, we train our model with less data and evaluate performance. We randomly select 10\%, 1\%, 0.1\%, 0.01\%  of the 736,167 time steps for training. The model achieves good performance even with 73 time steps for training. All trajectories are from the complex tensegrity setup with different parameters. 

\textcolor{black}{The proposed solution achieves very low error at a magnitude of $10^{-5}$, since it: 1) introduces a first-principles approach in learning physical parameters; 2) removes redundant data from regression; 3) operates -for now- over relatively clean data from simulation before moving to real-world data. }

\vspace{-.1in}
\subsection{Model Generalization Experiment}
\vspace{-.1in}
\begin{wrapfigure}{r}{0.5\textwidth}
    \vspace{-.3in}
    \centering
    \includegraphics[width=0.49\textwidth]{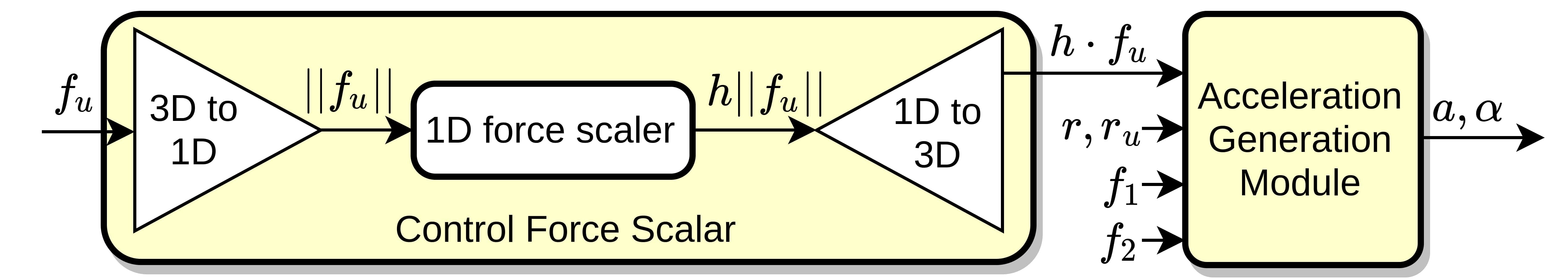}
    \vspace{-.1in}
    \caption{Control Force Scalar}
   \label{fig:control_force_scalar}
 \vspace{-0.2in}
\end{wrapfigure}
This section generalizes the physics engine trained with a dataset without external forces to a dataset with such forces. We are interested in evaluating: 1) how the physics engine performs for longer time horizons (e.g., after 2000 time steps); 2) if it can adapt to new scenarios. We generate a new dataset with 20,000 time steps trajectories with random directed perturbation forces $f_u$. The external force $f_u$ does not have the same scale as the internal spring forces, so we add a new scalar module with only one parameter $h$, as in Fig. \ref{fig:control_force_scalar}.  We also apply dimensionality reduction to improve data efficiency. The tuning process is to freeze all other modules' weights and train this by the new dataset. The error graphs are shown in Fig. \ref{fig:superball_diff_params_system_data_efficient_and_generalization} b).
\vspace{-.1in}
\section{Conclusion and Future Work}
\vspace{-.1in}
This paper proposes a differentiable physics engine for system identification of spring-rod systems based on first principles. The engine has three modules: force generation, acceleration generation and integration, which express the corresponding physical processes of spring-rod systems. This results in reduced data requirements and improved parameter accuracy. It also provides an explainable, accurate and fast physics engine. In the future, we plan to address contacts and friction. This will involve replacing the linear regressor with nonlinear models in the existing modules. To overcome noise in real data, we plan the addition of a residual network along with the nonlinear model. These changes may also help with temporal scalability. The ultimate mission, we generate policy from our identified engine and evaluate on the real platform to finally mitigate the reality gap.

\small
\bibliographystyle{plain}
\bibliography{references}

\begin{thebibliography}{10}

\bibitem{sain}
Anurag Ajay, Maria Bauza, Jiajun Wu, Nima Fazeli, Joshua~B Tenenbaum, Alberto
  Rodriguez, and Leslie~P Kaelbling.
\newblock {Combining Physical Simulators and Object-Based Networks for
  Control}.
\newblock In {\em IEEE International Conference on Robotics and Automation
  (ICRA)}, 2019.

\bibitem{asenov2019vid2param}
Martin Asenov, Michael Burke, Daniel Angelov, Todor Davchev, Kartic Subr, and
  Subramanian Ramamoorthy.
\newblock Vid2param: Online system identification from video for robotics
  applications.
\newblock {\em arXiv preprint arXiv:1907.06422}, 2019.

\bibitem{battaglia2016interaction}
Peter Battaglia, Razvan Pascanu, Matthew Lai, Danilo~Jimenez Rezende, et~al.
\newblock Interaction networks for learning about objects, relations and
  physics.
\newblock In {\em Advances in neural information processing systems}, pages
  4502--4510, 2016.

\bibitem{bruce2014superball}
Jonathan Bruce, Andrew~P Sabelhaus, Yangxin Chen, Dizhou Lu, Kyle Morse, Sophie
  Milam, Ken Caluwaerts, Alice~M Agogino, and Vytas SunSpiral.
\newblock Superball: Exploring tensegrities for planetary probes.
\newblock {\em 12th International Symposium on Artificial Intelligence,
  Robotics, and Automation in Space (i-SAIRAS)}, 2014.

\bibitem{chen2020design}
Muhao Chen, Jiacheng Liu, and Robert~E Skelton.
\newblock Design and control of tensegrity morphing airfoils.
\newblock {\em Mechanics Research Communications}, 103:103480, 2020.

\bibitem{de2018end}
Filipe de~Avila Belbute-Peres, Kevin Smith, Kelsey Allen, Josh Tenenbaum, and
  J~Zico Kolter.
\newblock End-to-end differentiable physics for learning and control.
\newblock In {\em Advances in Neural Information Processing Systems}, pages
  7178--7189, 2018.

\bibitem{golemo2018sim}
Florian Golemo, Adrien~Ali Taiga, Aaron Courville, and Pierre-Yves Oudeyer.
\newblock Sim-to-real transfer with neural-augmented robot simulation.
\newblock In {\em Conference on Robot Learning}, pages 817--828, 2018.

\bibitem{hansen2001completely}
Nikolaus Hansen and Andreas Ostermeier.
\newblock Completely derandomized self-adaptation in evolution strategies.
\newblock {\em Evolutionary computation}, 9(2):159--195, 2001.

\bibitem{heiden2019interactive}
Eric Heiden, David Millard, Hejia Zhang, and Gaurav~S Sukhatme.
\newblock Interactive differentiable simulation.
\newblock {\em arXiv preprint arXiv:1905.10706}, 2019.

\bibitem{hu2019difftaichi}
Yuanming Hu, Luke Anderson, Tzu-Mao Li, Qi~Sun, Nathan Carr, Jonathan
  Ragan-Kelley, and Fr{\'e}do Durand.
\newblock Difftaichi: Differentiable programming for physical simulation.
\newblock {\em arXiv preprint arXiv:1910.00935}, 2019.

\bibitem{hu2019chainqueen}
Yuanming Hu, Jiancheng Liu, Andrew Spielberg, Joshua~B Tenenbaum, William~T
  Freeman, Jiajun Wu, Daniela Rus, and Wojciech Matusik.
\newblock Chainqueen: A real-time differentiable physical simulator for soft
  robotics.
\newblock In {\em 2019 International Conference on Robotics and Automation
  (ICRA)}, pages 6265--6271. IEEE, 2019.

\bibitem{landry2019differentiable}
Benoit Landry, Zachary Manchester, and Marco Pavone.
\newblock A differentiable augmented lagrangian method for bilevel nonlinear
  optimization.
\newblock {\em arXiv preprint arXiv:1902.03319}, 2019.

\bibitem{lessard2016bio}
Steven Lessard, Dennis Castro, William Asper, Shaurya~Deep Chopra, Leya~Breanna
  Baltaxe-Admony, Mircea Teodorescu, Vytas SunSpiral, and Adrian Agogino.
\newblock A bio-inspired tensegrity manipulator with multi-dof, structurally
  compliant joints.
\newblock In {\em 2016 IEEE/RSJ International Conference on Intelligent Robots
  and Systems (IROS)}, pages 5515--5520. IEEE, 2016.

\bibitem{li2019propagation}
Yunzhu Li, Jiajun Wu, Jun-Yan Zhu, Joshua~B Tenenbaum, Antonio Torralba, and
  Russ Tedrake.
\newblock Propagation networks for model-based control under partial
  observation.
\newblock In {\em 2019 International Conference on Robotics and Automation
  (ICRA)}, pages 1205--1211. IEEE, 2019.

\bibitem{liang2019differentiable}
Junbang Liang, Ming~C. Lin, and Vladlen Koltun.
\newblock Differentiable cloth simulation for inverse problems.
\newblock In {\em Conference on Neural Information Processing Systems
  (NeurIPS)}, 2019.

\bibitem{littlefield2019kinodynamic}
Z.~Littlefield, D.~Surovik, M.~Vespignani, J.~Bruce, W.~Wang, and K.~E. Bekris.
\newblock Kinodynamic planning for spherical tensegrity locomotion with
  effective gait primitives.
\newblock {\em International Journal of Robotics Research (IJRR)}, accepted
  2019.

\bibitem{mrowca2018flexible}
Damian Mrowca, Chengxu Zhuang, Elias Wang, Nick Haber, Li~F Fei-Fei, Josh
  Tenenbaum, and Daniel~L Yamins.
\newblock Flexible neural representation for physics prediction.
\newblock In {\em Advances in Neural Information Processing Systems}, pages
  8799--8810, 2018.

\bibitem{paszke2019pytorch}
Adam Paszke, Sam Gross, Francisco Massa, Adam Lerer, James Bradbury, Gregory
  Chanan, Trevor Killeen, Zeming Lin, Natalia Gimelshein, Luca Antiga, et~al.
\newblock Pytorch: An imperative style, high-performance deep learning library.
\newblock In {\em Advances in Neural Information Processing Systems}, pages
  8024--8035, 2019.

\bibitem{rosenblatt1958perceptron}
Frank Rosenblatt.
\newblock The perceptron: a probabilistic model for information storage and
  organization in the brain.
\newblock {\em Psychological review}, 65(6):386, 1958.

\bibitem{rumelhart1986learning}
David~E Rumelhart, Geoffrey~E Hinton, and Ronald~J Williams.
\newblock Learning representations by back-propagating errors.
\newblock {\em nature}, 323(6088):533--536, 1986.

\bibitem{sabelhaus2018design}
Andrew~P Sabelhaus, Lara~Janse van Vuuren, Ankita Joshi, Edward Zhu, Hunter~J
  Garnier, Kimberly~A Sover, Jesus Navarro, Adrian~K Agogino, and Alice~M
  Agogino.
\newblock Design, simulation, and testing of a flexible actuated spine for
  quadruped robots.
\newblock {\em arXiv preprint arXiv:1804.06527}, 2018.

\bibitem{spnets2018}
C.~Schenck and D.~Fox.
\newblock Spnets: Differentiable fluid dynamics for deep neural networks.
\newblock In {\em Proceedings of the Second Conference on Robot Learning
  (CoRL)}, Zurich, Switzerland, 2018.

\bibitem{jaxmd2019}
Samuel~S. Schoenholz and Ekin~D. Cubuk.
\newblock Jax m.d.: End-to-end differentiable, hardware accelerated, molecular
  dynamics in pure python.
\newblock \url{https://github.com/google/jax-md},
  \url{https://arxiv.org/abs/1912.04232}, 2019.

\bibitem{stewart2000implicit}
David Stewart and Jeffrey~C Trinkle.
\newblock An implicit time-stepping scheme for rigid body dynamics with coulomb
  friction.
\newblock In {\em Proceedings 2000 ICRA. Millennium Conference. IEEE
  International Conference on Robotics and Automation. Symposia Proceedings
  (Cat. No. 00CH37065)}, volume~1, pages 162--169. IEEE, 2000.

\bibitem{surovid2018any}
D.~Surovik, J.~Bruce, K.~Wang, M.~Vespignani, and K.~E. Bekris.
\newblock Any-axis tensegrity rolling via bootstrapped learning and symmetry
  reduction.
\newblock In {\em International Symposium on Experimental Robotics (ISER)},
  Buenos Aires, Argentina, 11/2018 2018.

\bibitem{surovik2019adaptive}
D~Surovik, K~Wang, M~Vespignani, J~Bruce, and K~E Bekris.
\newblock {Adaptive Tensegrity Locomotion: Controlling a Compliant Icosahedron
  with Symmetry-Reduced Reinforcement Learning}.
\newblock {\em International Journal of Robotics Research (IJRR)}, 2019.

\bibitem{swevers1997optimal}
Jan Swevers, Chris Ganseman, D~Bilgin Tukel, Joris De~Schutter, and Hendrik
  Van~Brussel.
\newblock Optimal robot excitation and identification.
\newblock {\em IEEE transactions on robotics and automation}, 13(5):730--740,
  1997.

\end{thebibliography}
\end{document}